%% file: root.tex
\documentclass[letterpaper, 10 pt, conference]{ieeeconf}  
\usepackage{amsmath} 
\usepackage{amsfonts}

\usepackage{graphicx}
\usepackage{hyperref}       
\usepackage{cleveref}

\usepackage{booktabs}
\usepackage{multirow}
\usepackage[table]{xcolor}

\input{notation}

\IEEEoverridecommandlockouts                              

\usepackage{siunitx}

\title{\LARGE \bf
 Canonical Representation and Force-Based Pretraining of 3D Tactile for Dexterous Visuo-Tactile Policy Learning
}

\author{Tianhao Wu, Jinzhou Li$^{*}$, Jiyao Zhang$^{*}$, Mingdong Wu, Hao Dong
\thanks{Authors are with the Center on Frontiers of Computing Studies, School of Computer Science, Peking University, Beijing 100871, China, also with PKU-Agibot Lab, School of Computer Science, Peking University, Beijing 100871, China, and also with National Key Laboratory for Multimedia Information Processing, School of Computer Science, Peking University, Beijing 100871, China.}%
\thanks{* indicates equal contribution.}
\thanks{Corresponding to hao.dong@pku.edu.cn.}%
}

\def\figmk{Fig.~}
\def\tablemk{Tab.~}

\begin{document}

\maketitle
\thispagestyle{empty}
\pagestyle{empty}

\begin{abstract}

Tactile sensing plays a vital role in enabling robots to perform fine-grained, contact-rich tasks. However, the high dimensionality of tactile data, due to the large coverage on dexterous hands, poses significant challenges for effective tactile feature learning, especially for 3D tactile data, as there are no large standardized datasets and no strong pretrained backbones. To address these challenges, we propose a novel canonical representation that reduces the difficulty of 3D tactile feature learning and further introduces a force-based self-supervised pretraining task to capture both local and net force features, which are crucial for dexterous manipulation. Our method achieves an average success rate of 78\% across four fine-grained, contact-rich dexterous manipulation tasks in real-world experiments, demonstrating effectiveness and robustness compared to other methods. Further analysis shows that our method fully utilizes both spatial and force information from 3D tactile data to accomplish the tasks.
The codes and videos can be viewed at \url{https://3dtacdex.github.io}.

\end{abstract}

\section{INTRODUCTION}

Human hands are vital in daily life~\cite{li2022freedom}, enabling a wide range of tasks such as opening boxes and flipping objects. This level of dexterity is essential for integrating robots into everyday human activities. Vision-based imitation learning has shown great potential in teaching dexterous hands to perform various tasks~\cite{ze20243d, wang2024dexcap, guzey2023dexterity}. While simpler tasks like pick-and-place operations can achieve high success rates, more fine-grained and contact-rich tasks—such as flip object, remain significantly challenging. These tasks involve precise control of force, nuanced coordination of different fingers, and continuous feedback during manipulation. A key factor in successfully executing such tasks is tactile sensing~\cite{johansson1992somatosensory}.

To enable dexterous hands to perceive contact, current approaches typically equip them with tactile sensors. These sensors can be mainly categorized into vision-based tactile sensors, such as GelSight~\cite{patel2021digger}, DIGIT \cite{9018215}, and distributed tactile sensors, like uSkin~\cite{tomo2017covering}. 
Distributed tactile sensors are particularly well-suited for various robotic structures due to their small size, which allows for easy integration. Their robustness also makes them reliable in diverse environments, leading to widespread use in many systems~\cite{guzey2024see,bhirangi2023all,hu2023dexterous}. However, distributed tactile sensors typically have lots of taxels and cover large areas on dexterous hand~\cite{yang2023tacgnn}, leading to high-dimensional input. Moreover, different dexterous hands often use different types of distributed tactile sensors with varying sensor distributions, resulting in a lack of large-scale standardized datasets. This poses challenges in effectively learning tactile features for dexterous manipulation.

Considering the power of visual backbones, many works~\cite{funabashi2020stable, guzey2023dexterity} convert tactile data into 2D images to reduce the complexity of learning useful tactile features. However, this transformation leads to the change and loss of part spatial information between different taxels. To preserve these spatial relationships, most approaches represent 3D tactile data as a graph and use graph neural networks (GNNs)\cite{scarselli2008graph} to encode tactile signals\cite{yang2023tacgnn,funabashi2022multi}. However, these methods focus on specific tasks and require large data to learn effective features. Inspired by the success of the pretraining strategy in vision-based learning, T-DEX~\cite{guzey2023dexterity} collects tactile play data through interaction with various objects and pretrain tactile encoder with self-supervised learning. This pretraining improves the efficiency of feature learning and enhances diverse downstream robotic manipulation tasks. However, it still relies on 2D images as the tactile representation. As a result, efficiently learning 3D tactile data features for dexterous manipulation remains a challenge.

\begin{figure}[t!]
    \vspace{2mm}
    \centering
    \includegraphics[width=0.9\linewidth]{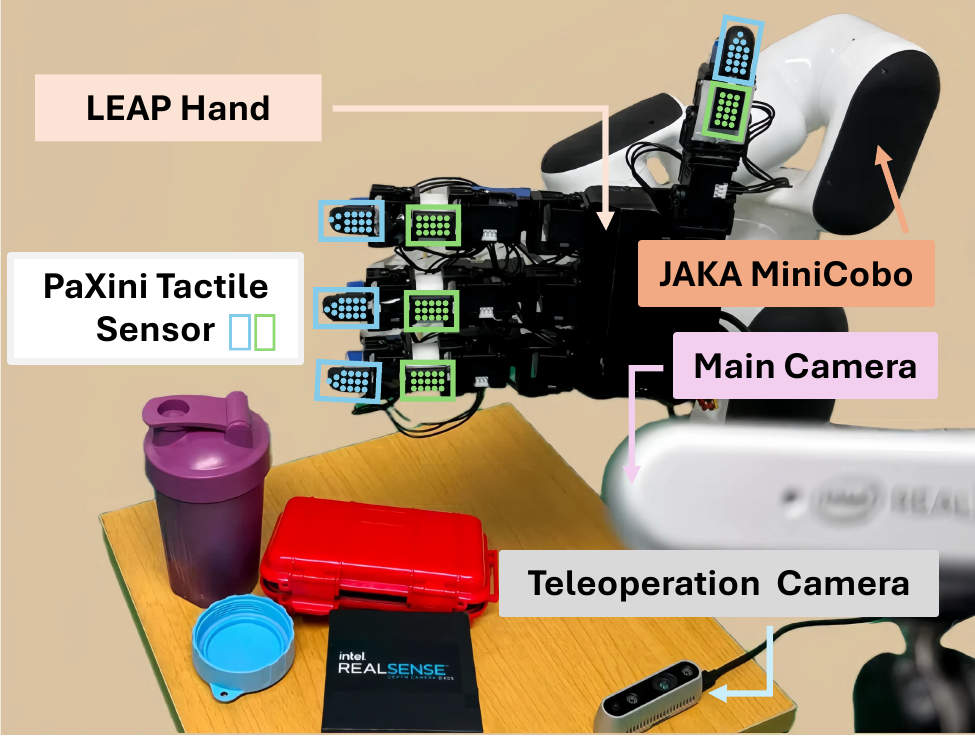}
    \caption{\textbf{Real Robot System.} Our system uses one camera and distributed tactile sensors to achieve dexterous, fine-grained, contact-rich tasks. The teleoperation camera is only used for data collection, not policy learning.}
    \label{fig:system}
    \vspace{-5mm}
\end{figure}

To address the difficulty of 3D tactile feature learning, we first propose a novel canonical representation of 3D tactile data, which canonicalizes the coordinates of taxels in each sensor into a unified frame. This canonicalization aligns the features of differently distributed sensors and reduces the feature space. Additionally, it amplifies the distances between taxels within the same sensor, facilitating the capture of more localized features. We further propose a force-based, self-supervised prediction task for pretraining 3D tactile data, given the importance of force usage in object manipulation. The pretraining tasks include both masked local force prediction and net force prediction, encouraging the encoder to learn features related to both local and net force relationships.

To demonstrate the effectiveness of our method, we integrate the pretrained tactile encoder into an imitation learning framework and evaluate it in the real world on four fine-grained, contact-rich tasks: open box, reorientation, flip, and assembly. Comparative results demonstrate the effectiveness of our method compared to other baselines. Ablation studies confirm the importance of our proposed canonical representation and force-based pretraining. Additionally, our analysis shows that the policy effectively utilizes both the spatial and force information from the 3D tactile data.

In summary, our contributions are as follows: (i) We propose a novel canonical representation for 3D tactile data that effectively improves 3D tactile data feature learning. (ii) We propose a novel force-based self-supervised pertaining task on tactile play data, including local force and net force prediction, enhancing downstream dexterous manipulation policy learning. (iii) We demonstrate the effectiveness and robustness of our method through a range of real-world experiments using a dexterous hand.

\section{RELATED WORK}

\subsection{Tactile for Dexterous Manipulation}

Tactile sensing has been widely used to enhance dexterous manipulation. Current approaches primarily utilize either vision-based tactile sensors~\cite{patel2021digger} or distributed tactile sensors~\cite{tomo2017covering}. Vision-based tactile sensors are typically only mounted on the fingertips of dexterous hands~\cite{qi2023general}, due to their large size. In contrast, distributed tactile sensors can cover a larger area~\cite{bhirangi2023all}. Thus, we choose to use distributed tactile sensors in this work. 

There are two main approaches to learning manipulation policies with distributed tactile sensors. One is simulation-to-reality, where simulation can generate large amounts of tactile data, making the learning process more efficient. However, there is a significant gap between tactile data in simulation and the real world. To close the sim-to-real gap, most works use discrete tactile signals~\cite{yuan2024robot, yin2023rotating, yin2024learning} or only activated tactile positions~\cite{yang2023tacgnn} as input, limiting the full potential of tactile sensors. The other approach is learning directly in the real world. Due to the large coverage area of distributed sensors, tactile input can be high-dimensional, especially for dexterous hands, posing challenges for efficient learning. Pretraining with play data has been proposed to improve efficiency~\cite{guzey2023dexterity, guzey2024see}, but these works rely only on 2D tactile images. In contrast, we propose a pretraining strategy and representation specifically for 3D tactile data.

\subsection{Tactile Representaiton}

Different tactile representations convey various types of information and can be encoded using different methods. For low-dimensional tactile data, directly applying MLP to flattened tactile readings~\cite{lin2024learning} or converted 3D vectors~\cite{zambelli2021learning} can capture useful tactile features. However, as sensor coverage increases, the dimensionality of the data grows significantly, making direct encoding of tactile readings inefficient. To leverage powerful visual backbones, some methods convert raw tactile readings into RGB images by mapping the tri-axis forces to three channels~\cite{sundaram2019learning, funabashi2020stable, sferrazza2023power}, using visual backbones for encoding. However, such 2D information changes the inherent spatial relationship of taxels in the same sensor and does not contain the spatial relationship of taxels in different sensors that are distributed on the different parts of the robot. To preserve spatial relationships, graph-based methods have been applied to tactile data by treating each taxel as a node, connecting them with either predefined~\cite{funabashi2022multi} or dynamically changing graph~\cite{yang2023tacgnn}. Nevertheless, the representation in these works only use a subset of the available tactile information in 3D space, such as the 3D position~\cite{yang2023tacgnn} or 3D force of the taxels~\cite{funabashi2022multi}. Our work fully leverages both the 6D pose and 3D force of each taxel, and we propose a novel canonical representation to more effectively learn features from such complex tactile data.

\subsection{Tactile Pretraining}

Due to the high-dimensional of tactile data, pretraining is an effective strategy for improving the efficiency of downstream task learning. Different pretraining strategies encourage the encoder to learn distinct features. Aligning vision and tactile data has been widely studied for pretraining to understand relationships between different data modalities~\cite{li2019connecting, zambelli2021learning, yuan2017connecting, luo2018vitac}. However, these approaches primarily focus on inter-modal pretraining for multi-modal learning. Our work focuses on intra-modal pretraining.

A common approach for intra-modal pretraining typically involves augmenting the data and encouraging the encoder to match the augmented data with the original~\cite{grill2020bootstrap}, enhancing the encoder's ability to discriminate between different data patterns. However, most powerful intra-modal pretraining methods are designed for 2D images, which require representing tactile data as 2D images~\cite{guzey2023dexterity, dave2024multimodal}, failing to fully leverage the spatial information in 3D tactile data. In contrast, we focus on pretraining for 3D tactile data, and instead of enhancing the encoder’s discriminative ability, we encourage it to learn features related to force.

\section{ROBOT SYSTEM SETUP}

As shown in \figmk\ref{fig:system}, our system consists of a 6-Dof JAKA MiniCobo robot arm and a 16-Dof Leap Hand~\cite{shaw2023leaphand} dexterous hand with four fingers. The Leap Hand is equipped with PaXini tactile sensors, Each finger has two types of sensors: one for the fingertip and another for the fingerpad. Both types of sensors have a 3x5 array of taxels, but taxel distribution is slightly different. Each taxel measuring tri-axial forces $\force \in \mathbb{R}^3$. A single Intel RealSense D415 camera is mounted diagonally of the robot to capture visual information.

For expert demonstration collection, we use an additional Intel RealSense D415 camera with HaMeR~\cite{pavlakos2024reconstructing} to track human hand pose, use Dexpilot~\cite{handa2020dexpilot} to retarget and teleoperate the robot. The robot arm is controlled with a target end-effector pose consisting of 3-Dof translation and 4-Dof quaternion, while the robot hand is controlled with 16-Dof target hand joint positions. Both demonstration collection and inference are performed at a frequency of 5 Hz.

\section{METHOD}

\begin{figure*}[t!]
    \centering
    \vspace{5mm}
    \includegraphics[trim=1 0 2 0, clip, width=\linewidth]{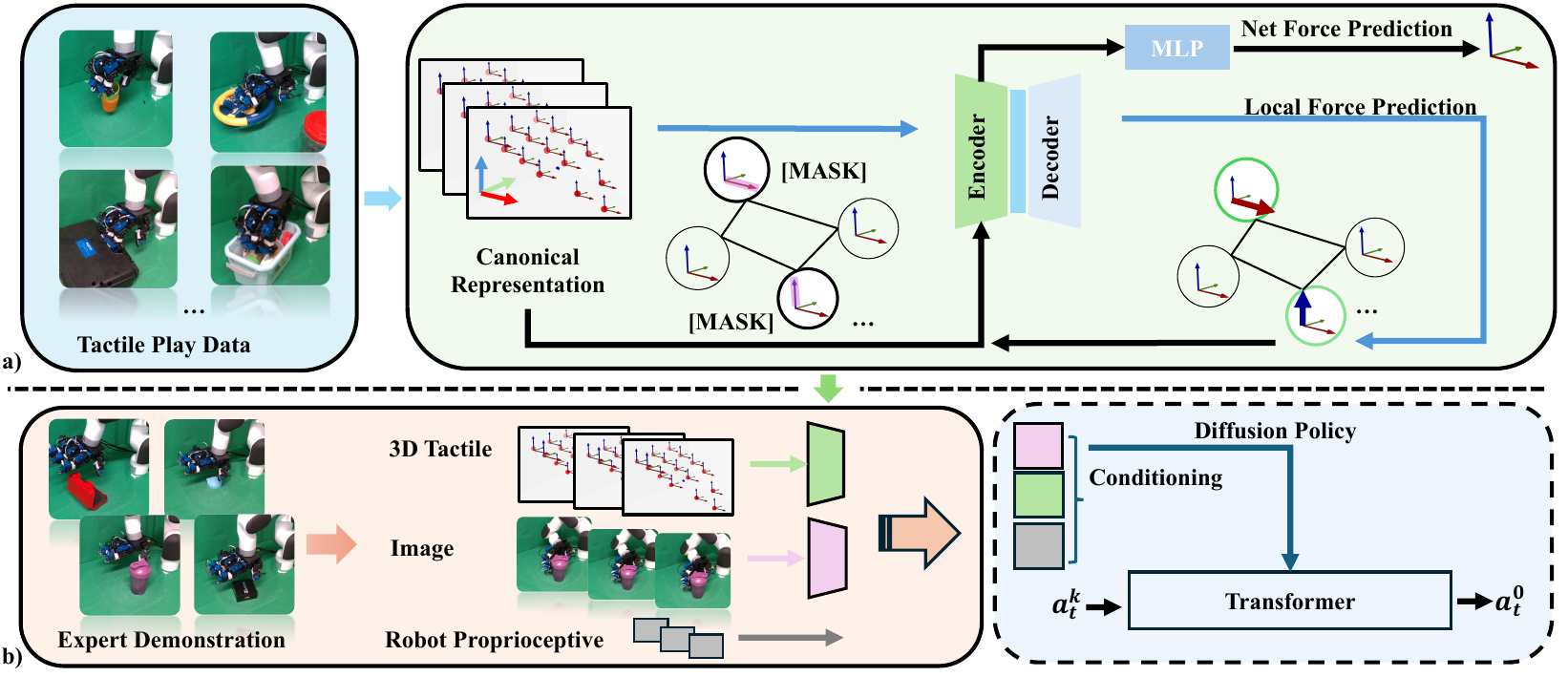}
    \caption{\textbf{Pipeline.} a) Pretraining on play data with our canonical representation and force-based task. Local force prediction: a portion of the tactile force is randomly masked, encoded into a latent representation, and then decoded to predict the masked forces. Net force prediction: the predicted masked forces are substituted back into the original data and encoded again to predict the net force. The local force prediction and net force prediction share the same encoder. b) Incorporating the pretrained encoder within the imitation learning framework for downstream dexterous manipulation.}
    \label{fig:pipeline}
    \vspace{-5mm}
\end{figure*}

We focus on the problem of leveraging 3D tactile data from distributed tactile sensors for learning visuo-tactile dexterous manipulation policies. To reduce the difficulty of learning features from complex 3D tactile data, we canonicalize the data into a unit frame, in \ref{sec:canoical-rep}. We then pretrain the tactile encoder using self-supervised force-based prediction tasks to enhance local and net force feature learning, in \ref{sec:force-predict}. This pretrained tactile encoder is subsequently used for visuo-tactile policy learning, in \ref{sec:policy-learning}.

\subsection{Canonical Tactile Representation}\label{sec:canoical-rep}

\begin{figure}[t!]
    \centering
    \includegraphics[width=0.8\linewidth]{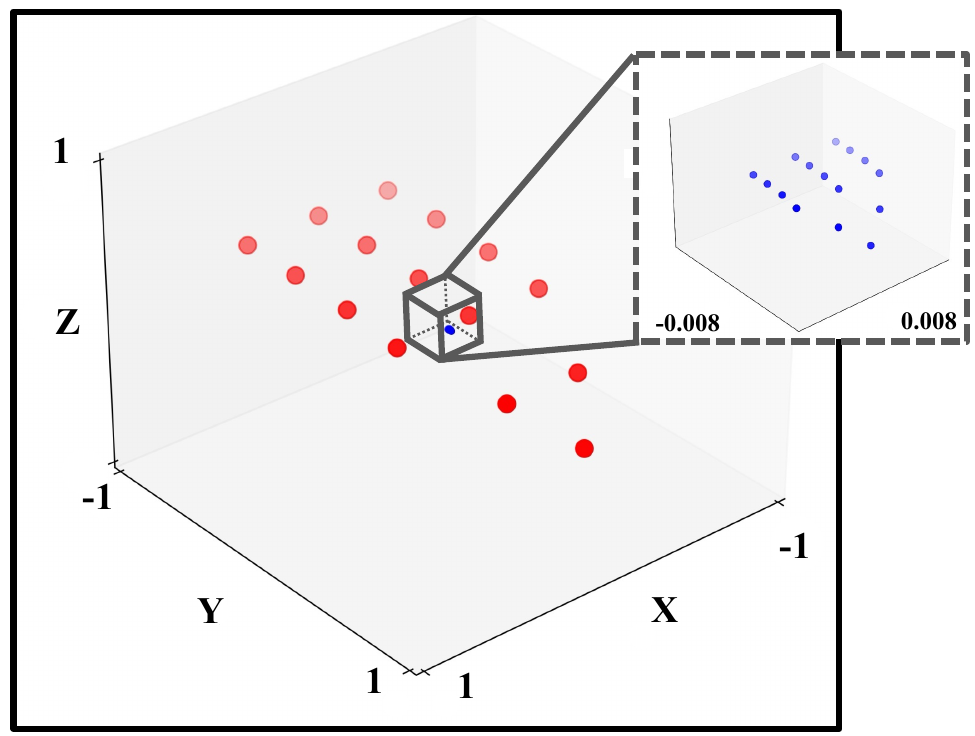}
    \caption{\textbf{Comparison of Canonical Representation and Original Representation.} We visualize the coordinates of each taxel in the fingertip sensor before and after canonicalization. The blue dots represent the original taxel coordinates, which are difficult to distinguish when input to the encoder. In contrast, the red dots represent the taxel coordinates after canonicalization. With canonicalization, the coordinates of each taxel become more discriminative, and sensors of the same type have a consistent representation, reducing the feature space. Canonicalization with diagonal length maintains the inherent spatial relationships between taxels on each sensor pad, the same as the original representation.}
    \label{fig:canonical tactile rep}
    \vspace{-5mm}
\end{figure}
To preserve the spatial relationships of each taxel, we aim to use 3D tactile data instead of converting it into a 2D image. For each taxel of the distributed tactile sensor, in addition to the 3D force $\force$, we can also obtain the 6D pose $\pose \in \mathbb{R}^6$ by computing forward kinematics. This information shows how the force is applied at every step. However, since a large number of taxels are distributed across different parts of the fingers, using this 9D tactile representation results in a vast feature space, making it difficult to learn meaningful tactile features. Additionally, though the taxels within a sensor are distributed sparsely, the distances between them are very small (e.g., less than 4 millimeters), making it challenging to capture local features within the same sensor.

To address the challenges, we propose to canonicalize the 9D tactile representation. Specifically, we normalize each taxel's coordinate within the same sensor into a unit frame (ranging from -1 to 1 for each axis) by computing the diagonal length of the original coordinates within the sensor frame. As shown in \figmk\ref{fig:canonical tactile rep}, the 3D position of each taxel in the unit frame is denoted as $\trans \in \mathbb{R}^3$. However, this representation only captures the spatial relationships between taxels within the same sensor, without accounting for the spatial relationships between different sensors. Therefore, we also include the 6D pose of each sensor’s origin with respect to the hand's base, denoted as $\pose^s$, into the representation. As a result, the representation for each taxel is represented as $\rep = [\pose^s, \trans, \force]$. Although this representation has a higher dimension, it effectively reduces the feature space because the features of different sensors become more aligned due to the canonicalized coordinates. Additionally, this canonicalization amplifies the relative distance between taxels within the same sensor, making their features more distinguishable for the neural network. This facilitates the capture of more localized features for each taxel.

However, this representation still suffers from the inherent sparsity of the distributed tactile sensor. To address this, we utilize a graph neural network~\cite{hou2022graphmae} to encode our proposed representation. We define the tactile information as a set of our proposed 12D representations, e.g., $\set = \{\rep_1, \rep_2,...,\rep_n \}$. Based on $\set$, we construct the graph $\graph = (\set,\edge)$, where $\edge$ represents the edges defined by the 4-neighbourhood of each tactile node.

\subsection{Force-based Pretraining}\label{sec:force-predict}

While the canonical tactile representation can ease the difficulty of tactile feature learning, it does not ensure that the neural network will learn the features essential for manipulation and can be low-efficiency if trained on specific tasks only. Pretraining, however, can encourage the encoder to learn the inherent structures of the data~\cite{dave2024multimodal} and improve the efficiency of learning for downstream tasks~\cite{guzey2023dexterity}.

What kind of pretraining can we use for 3D tactile data to improve dexterous manipulation policies? When humans manipulate objects, we carefully apply force to achieve the desired object pose. Inspired by this, we propose pretraining the 3D tactile data based on force. When applying force, it is essential to consider how each finger part applies local force so that the net force moves the object as intended. Consequently, we designed two force-based self-supervised pretraining tasks: the first predicts the local force, and the second predicts the net force. Since our robot system differs from T-DEX~\cite{guzey2023dexterity}, we follow their method to collect our own play data for pretraining. We use GraphMAE~\cite{hou2022graphmae} as the backbone for pretraining.

\paragraph*{Local Force Prediction}

To help the encoder learn the local features of each taxel, we design a masked force prediction task. Since the force applied to each taxel can propagate to its neighboring, we randomly mask part of the tactile force and use the masked tactile data as input for the encoder. The encoder first encodes the tactile data, then decodes the latent representation to reconstruct the original tactile data, as shown in \figmk\ref{fig:pipeline}. We compute MSE loss between the reconstructed and original force values, only for the masked forces. This pretraining approach helps the encoder learn the relationships between local forces.

\paragraph*{Net Force Prediction}

To help the encoder understand the relationship between local and net forces, we design a self-supervised task for net force prediction. Given the original 3D tactile data, we compute the net force $\force^n_G$ based on each taxel's pose and force. This $\force^n_G$ serves as the target for prediction. As shown in \figmk\ref{fig:pipeline}, after predicting the local force, we substitute the predicted force values into the original tactile data and use the same encoder to encode this modified data into a latent representation. We then use an MLP to predict the net force $\force^n_P$. The MSE loss is calculated between $\force^n_G$ and $\force^n_P$. By further predicting the net force, the encoder learns to capture the feature of both local and net force, which benefits downstream tasks.

\subsection{Visuo-Tactile Policy Learning}\label{sec:policy-learning}

After pretraining the tactile encoder, we use imitation learning to learn visuo-tactile policy for dexterous manipulation. Given the diffusion model's ability to model complex action distributions~\cite{wu2022targf}, we adopt the diffusion policy~\cite{chi2023diffusionpolicy} as our backbone. We replace the vision backbone with DinoV2~\cite{oquab2023dinov2} for better visual feature extraction and integrate tactile data as an additional input, encoded by the pretrained tactile encoder. The tactile features are concatenated with visual features for policy learning. We fine-tune the encoder during downstream tasks training, following diffusion policy~\cite{chi2023diffusionpolicy}. We also include the robot's proprioceptive state, including 3D position, 4D quaternion of the arm, and 16D joint position of the hand. Our action space is target actions rather than states, as these actions implicitly capture force usage, crucial for accomplishing diverse tasks~\cite{zhao2023learning,yangace}.



\section{EXPERIMENTS}

\begin{figure*}[h]
    \centering
    \vspace{5mm}
    \includegraphics[trim=0 0 2 0, clip, width=\linewidth]{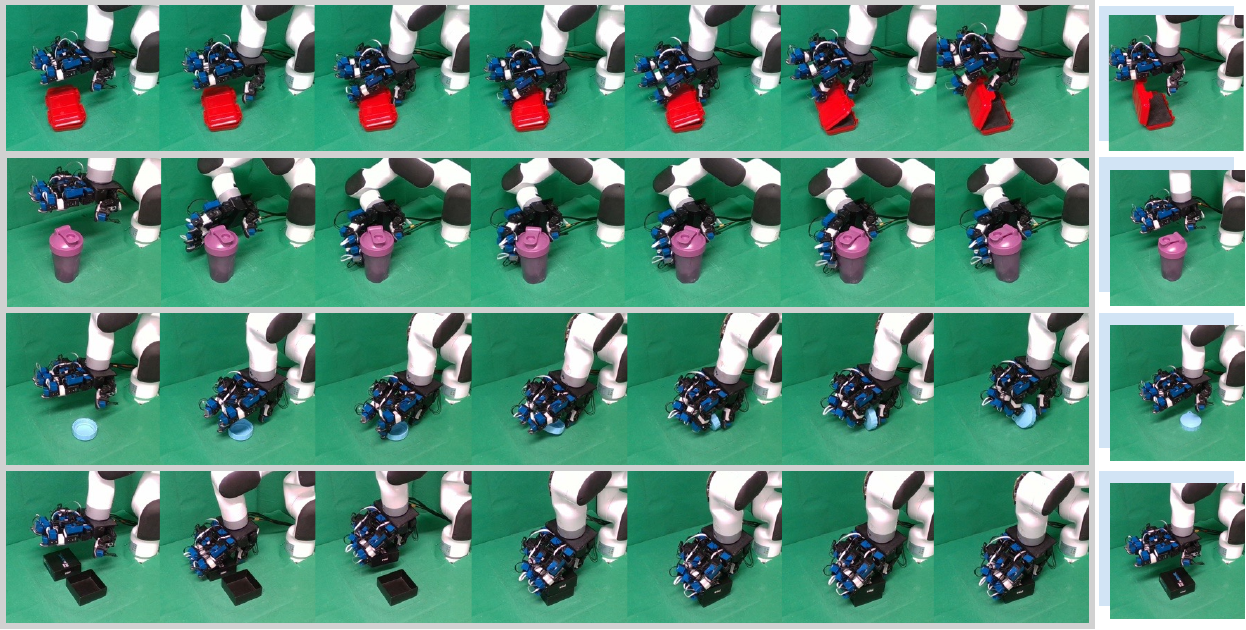}
    \caption{\textbf{Visualization of Our Policy's Rollout on Four Fine-Grained, Contact-Rich Tasks.} Note this is the view of the robot's observation.}
    \label{fig:tasks}
    \vspace{-5mm}
\end{figure*}

We conduct comprehensive real-world experiments to validate the following questions:
\begin{itemize} 
    \item Can our canonical tactile representation help in learning features from complex 3D tactile data? 
    \item Does our force-based pretraining improve visuo-tactile policy performance? 
    \item What role do spatial and force information of tactile data play during dexterous manipulation?
 \end{itemize}

\subsection{Dexteous Manipulation Tasks}

We conduct experiments on four dexterous fine-grained, contact-rich manipulation tasks, as shown in \figmk\ref{fig:tasks}. Each experiment run will be limited to a maximum of 600 steps. Each method will be evaluated on each task of 10 experiment runs. 1) Open Box: This task requires the robot to open a box using the thumb and index finger. The robot needs to first reach the box, grasp the upper part, and then carefully adjust its finger to open the box without pushing it. The challenge is maintaining a firm hold on the upper part during opening to prevent it from loosening and falling. The box is placed randomly within an 18x12 cm area for each run. Success is achieved if the upper part of the box stays in place after opening. 2) Reorientation: This task requires the robot to continuously reorient a bottle until it points in a specific direction. The robot needs to reach the bottle and coordinate its four fingers to reorient it without pushing it down. The challenge is the precise coordination of the fingers, and the task is long-horizon. The bottle is placed in a random pose within an 18x12 cm area for each run. Success is achieved if the bottle is within 10 degrees of the target direction. 3) Flip: This task requires the robot to flip a bottle cap using the thumb, middle, and index finger. The robot needs to reach the cap, grasp it, lift one side of the cap, and use the index finger to flip the cap. The challenge involves precise finger coordination and force application, with severe occlusion and ambiguity during the process. The cap is placed in a fixed position with random orientations, and success is achieved if the cap is flipped by 180 degrees. 4) Assembly: This task requires the robot to grasp one part of a box and assemble it with another. The robot needs to reach, grasp, move, and gradually insert one part into the other. The challenge is making fine adjustments based on feedback while handling high occlusion and ambiguity. The box parts are in fixed positions, and success is achieved when one part is successfully inserted into the other.

\subsection{Baselines}
We compare our method with the following baselines, which all use the same visual backbone, diffusion policy backbone, visual observation, robot proprioceptive state, and action space as ours, but with different types of tactile representation and pertaining.
1) DP: We implement the diffusion policy without using tactile data or pertaining for this baseline.
2) HATO: HATO~\cite{lin2024learning} uses MLP to encode the tactile. We flatten force values and use MLP to encode tactile data for this baseline.
3) T-DEX: T-DEX~\cite{guzey2023dexterity} convert the raw tactile into 2D image, and pretrain with BYOL~\cite{grill2020bootstrap}. For this baseline, we follow their procedure, first converting raw tactile into 2D images, then pretrain encoder on our own collected dataset, then encoder for diffusion policy learning.
4) GNN: We use the 9D tactile representation (e.g., the 6D pose of each taxel with 3D tactile force values) as input for this baseline, use graph attention networks~\cite{velivckovic2018graph} to encode tactile, which is the same GNN backbone as ours. Since there are no pertaining strategy designed especially for 9D tactile representation, We do not pretrain for this baseline.

\subsection{Manipulation Policy Comparsion}

\begin{table}[h]
    \caption{\textbf{Success Rate of Different Manipulation Policies.}}
    \vspace{-5mm}
    \label{table:main}
    \begin{center}
        \resizebox{1\linewidth}{!}{
            \input{Tables/main.tex}

        }
    \end{center}
    \vspace{-5mm}
\end{table}

As shown in \tablemk\ref{table:main}, our approach achieves the highest success rate across all tasks. Most baselines perform well on the open box and reorientation tasks but struggle with the assembly and flip tasks. Interestingly, we found that even without tactile feedback, \textit{DP} still achieves a high success rate on open box and reorientation tasks. This is mainly because these tasks do not involve significant occlusion or ambiguity during manipulation, allowing \textit{DP} to successfully find and manipulate objects using visual input and robot state alone. In contrast, even with rich tactile information, \textit{GNN} consistently fails across all tasks, we observe that the finger or the hand usually shakes during the manipulation, preventing it from finishing the task. Compared to \textit{GNN}, \textit{HATO}, which only uses tactile force values, is able to accomplish some tasks, demonstrating the difficulty of learning spatial and force information from 3D tactile data simultaneously. \textit{T-DEX} performs better than the other baselines, showing that even 2D tactile data with pretraining can achieve high success rates, though it struggles with the flip task.

The flip task requires extremely precise coordination between the fingers and relies on tactile feedback to ensure a firm grasp and accurate force application. For this task, we observed that \textit{DP} hesitates to grasp the bottle cap and often reaches the maximum number of steps without succeeding, mainly due to the lack of tactile feedback. While \textit{HATO} can reach the object accurately, it usually does not perform grasp or lift. \textit{T-DEX} fails primarily due to an unstable initial grasp, which leads to difficulties during middle-finger lifting and index-finger reorientation. This underscores the importance of the spatial information provided by 3D tactile data. 

\subsection{Importance of Representation and Pretraining}

\begin{table}[h]
    \caption{\textbf{Success Rate of Ablation.} CR: our proposed canonical representation. PRE: our proposed force-based pertaining. }
    \vspace{-5mm}
    \label{table:method-ablation}
    \begin{center}
        \resizebox{1\linewidth}{!}{
            \input{Tables/method_ablation.tex}
        }
    \end{center}
    \vspace{-5mm}
\end{table}

To validate the effectiveness of our canonical representation and force-based pretraining, we conduct ablation studies across all tasks. As shown in \tablemk\ref{table:method-ablation}, using canonical representation achieves 48\% success rate, even without pretraining. However, without canonical representation, even with pretraining, the policy fails to complete any task. We observed that the policy moves to a specific hand joint position upon starting inference and then repeats similar actions. Analyzing the output of tactile encoder, we found that without canonical representation, the encoder outputs similar features for taxels within the same sensor pad, failing to perceive fine-grained differences. This validates the necessity of canonical representation. Based on such representation, pretraining further enables the encoder to learn more useful features, increasing the success rate to 78\%.

\subsection{Effect of Force-based Pretraining Tasks}

\begin{table}[h]
    \caption{\textbf{Success Rate of Pertaining Task.} NF: net force prediction. LF: masked local force prediction. }
    \vspace{-5mm}
    \label{table:pretrain-ablation}
    \begin{center}
        \resizebox{1\linewidth}{!}{
            \input{Tables/pretrain_task_ablation}
        }
    \end{center}
    \vspace{-5mm}
\end{table}

To validate the effectiveness of our pretraining tasks, we conducted experiments using only one pretraining task at a time. As shown in \tablemk\ref{table:pretrain-ablation}, omitting either pertaining task leads the encoder to focus solely on either local or net force features, resulting in a significant performance drop in all tasks. The results also show that net force prediction is more critical for achieving the tasks.

\subsection{Role of Spatial Information and Force Information}

We conducted an ablation study to validate the use of spatial and force information in our policy. In the spatial ablation, the tactile sensor's 6D pose was fixed at the initial state, while in the force ablation, all tactile forces were set to zero. In both cases, the robot failed to flip the cap at all. In the spatial ablation, once the robot reached the object and attempted to grasp it, the thumb oscillated randomly, preventing further manipulation. In the force ablation, although the robot reached the object and attempted to grasp it, it consistently failed due to an unstable grasp or continuous adjustments. These results demonstrate that our policy leverages spatial information for forming gross hand poses and force information for more fine-grained adjustments.

\subsection{Generlization to Unseen Objects}

\begin{figure}[t!]
    \centering
    \vspace{3mm}
    \includegraphics[trim=0 0 2 6, clip, width=\linewidth]{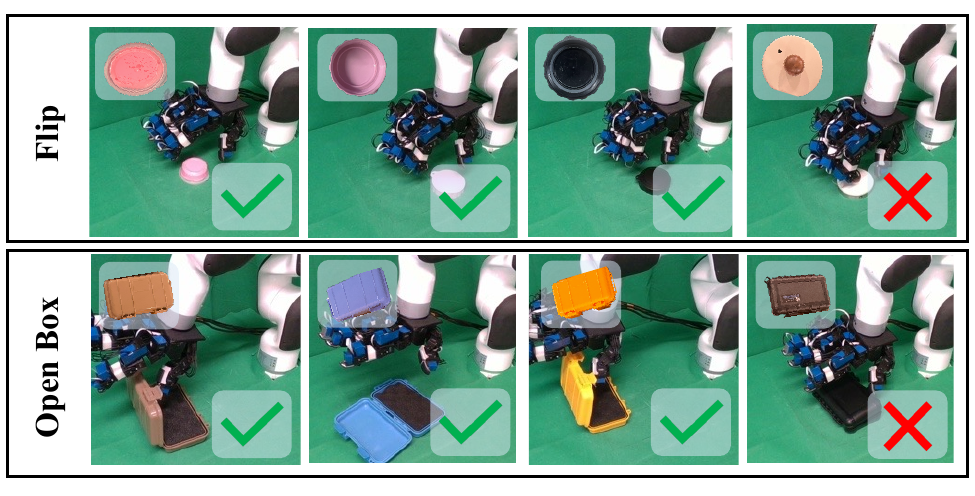}
    \vspace{-5mm}
    \caption{\textbf{Visualization of Our Policy on Unseen Objects.}}
    \label{fig:gen}
    \vspace{-5mm}
\end{figure}

To validate the generalization of our method, we tested the policy with four unseen objects exhibiting diverse color, geometry, and dynamics, with each object being tested twice for the open box and flip tasks. As shown in \figmk\ref{fig:gen}, our policy successfully opens the box 5 times out of 8 tries. For one failure of the open box task, although the hand opened the box to a certain degree that generally won't fall down, it fell down due to completely different friction properties of the box. For the flip task, the policy succeeded 6 times out of 8 tries, demonstrating the generalization ability of our method.

\section{CONCLUSIONS}

In this work, we enhance 3D tactile feature learning by proposing a novel canonical representation that aligns differently distributed tactile sensor readings, reduces the feature space, and increases the discriminability of each taxel within the same sensor. We also introduce a force-based self-supervised pretraining task to encourage using both spatial and force information. Real-world experiments using the pretrained encoder for downstream dexterous, fine-grained, contact-rich tasks demonstrate the effectiveness and robustness of our methods.

\textbf{Limitations and Future Work.}  Our policy shows limited generalization when encountering objects with significantly different shapes and dynamics. Quick adaptation using tactile could be a direction for future work.


\section*{ACKNOWLEDGMENT}
This work is supported by the National Youth Talent Support Program (8200800081) and National Natural Science Foundation of China (No. 62376006).

\addtolength{\textheight}{-0.5cm}   


\bibliographystyle{IEEEtran}
\bibliography{ref}

\end{document}

%% file: notation.tex
\newcommand{\force}{\mathbf{F}}
\newcommand{\pose}{\mathbf{P}}
\newcommand{\trans}{\mathbf{T}}

\newcommand{\rep}{\mathbf{R}}
\newcommand{\set}{\mathbf{S}}

\newcommand{\graph}{\mathbf{G}}
\newcommand{\edge}{\mathbf{E}}

%% file: Tables/main.tex
\begin{tabular}{lccccc}

\toprule
\textbf{Method}                 & \textbf{Open Box} & \textbf{Reorientation} & \textbf{Flip}     & \textbf{Assembly}    & \textbf{Avg} \\ 
\midrule
\textbf{DP}                     & 90\%              & 60\%                   & 20\%              & 40\%                   & 53\%         \\
\textbf{HATO}                   & 70\%              & 60\%                   & 10\%              & 50\%                   & 48\%         \\
\textbf{T-DEX}                  & 80\%              & 70\%                   & 40\%              & 60\%                   & 63\%         \\
\textbf{GNN}                    &  0\%              &  0\%                   &  0\%              & 0\%                   &  0\%        \\
\rowcolor{gray!20}\textbf{Ours} & \textbf{90\%}     & \textbf{70\%}          & \textbf{80\%}     & \textbf{70\%}        & \textbf{78\%}  \\ 
\bottomrule
\end{tabular}

%% file: Tables/method_ablation.tex
\begin{tabular}{lccccc}
\toprule
\textbf{Method}                 & \textbf{Open Box} & \textbf{Reorientation} & \textbf{Flip}     & \textbf{Assembly} & \textbf{Avg} \\ 
\midrule
\textbf{Ours w/o CR \& PRE}     &  0\%              &  0\%                   &  0\%              &  0\%              &  0\%         \\
\textbf{Ours w/o CR}            &  0\%             &  0\%                  &  0\%             &  0\%             &  0\%        \\
\textbf{Ours w/o PRE}           &  60\%             &  60\%                  &  50\%             &  20\%             &  48\%        \\
\rowcolor{gray!20}\textbf{Ours} & \textbf{90\%}     & \textbf{70\%}          & \textbf{80\%}     & \textbf{70\%}     & \textbf{78\%}  \\ 
\bottomrule
\end{tabular}

%% file: Tables/pretrain_task_ablation.tex
\begin{tabular}{lccccc}
\toprule
\textbf{Method}                 & \textbf{Open Box} & \textbf{Reorientation} & \textbf{Flip}  & \textbf{Assembly}     & \textbf{Avg} \\ 
\midrule
\textbf{Ours w/o NF}            &  30\%             & 30\%                   &  40\%          &  10\%                 &  28\%         \\
\textbf{Ours w/o LF}            &  70\%             & 50\%                   &  30\%          &  40\%                 &  48\%        \\
\rowcolor{gray!20}\textbf{Ours} & \textbf{90\%}     & \textbf{70\%}          & \textbf{80\%}  & \textbf{70\%}         & \textbf{78\%}  \\ 
\bottomrule
\end{tabular}